\documentclass[letterpaper, 10 pt, conference]{ieeeconf}  

\IEEEoverridecommandlockouts                              
\usepackage{graphicx}
\usepackage{siunitx}
\usepackage[utf8]{inputenc}
\usepackage[T1]{fontenc}
\usepackage{amsfonts}
\newcommand{\fig}[1]{Fig.~\ref{#1}}

\usepackage{epsfig} 
\usepackage{amsmath} 
\usepackage{subcaption}
\usepackage{caption}
\usepackage{color}
\usepackage{verbatim}
\usepackage{wasysym}
\usepackage[table,xcdraw]{xcolor}
\usepackage{graphicx}
\usepackage{gensymb}
\usepackage{xcolor}
\usepackage{latexsym}
\usepackage{multirow}
\usepackage{amsmath}
\usepackage{booktabs,caption,fixltx2e}
\usepackage[flushleft]{threeparttable}
\usepackage[table,xcdraw]{xcolor}

\newcommand{\RNum}[1]{\uppercase\expandafter{\romannumeral #1\relax}}
\makeatletter
\newlength\tmp@\newlength\t@mp
\newcommand{\comp}[3]
  {\mathop{ \settowidth\tmp@{$\displaystyle\mathop{#1}^{#3}_{#2}$}
  \hbox to \tmp@{\hss \settowidth\t@mp{$\displaystyle #1$}\setlength\t@mp{.45\t@mp}
  $\displaystyle\mathop{#1}^{\hspace\t@mp #3}_{\hspace{-\t@mp}#2}$
  \hss} }}
\makeatother
\title{\LARGE \bf
SCALER: A Tough Versatile Quadruped Free-Climber Robot
}

\author{Yusuke Tanaka$^{\dagger}$, Yuki Shirai$^{\dagger*}$, Xuan Lin$^{\dagger*}$, Alexander Schperberg$^{\dagger}$, \\ Hayato Kato$^{\dagger}$, Alexander Swerdlow$^{\dagger}$, Naoya Kumagai$^{\dagger}$, and Dennis Hong$^{\dagger}$
\thanks{$^{\dagger}$ All authors are with the Department of Mechanical and Aerospace Engineering, University of California, Los Angeles, CA, USA 90095 {\tt\small \{yusuketanaka, yukishirai4869, maynight, aschperberg28, hayatokato, aswerdlow, kumanao1999, dennishong\}@g.ucla.edu.} $^*$Y. Shirai and X. Lin assert joint second authorship.}}

\begin{document}
\maketitle
\thispagestyle{empty}
\pagestyle{empty}

\begin{abstract}
This paper introduces SCALER, a quadrupedal robot that demonstrates climbing on bouldering walls, overhangs, ceilings and trotting on the ground. SCALER is one of the first high-degrees of freedom four-limbed robots that can free-climb under the Earth's gravity and one of the most mechanically efficient quadrupeds on the ground. Where other state-of-the-art climbers specialize in climbing, SCALER promises practical free-climbing with payload \textit{and} ground locomotion, which realizes true versatile mobility. A new climbing gait, SKATE gait, increases the payload by utilizing the SCALER body linkage mechanism.
SCALER achieves a maximum normalized locomotion speed of $1.87$ /s, or $0.56$ m/s on the ground and $1.0$ /min, or $0.35$ m/min in bouldering wall climbing. Payload capacity reaches $233$ \% of the SCALER weight on the ground and $35$ \% on the vertical wall. 
Our GOAT gripper, a mechanically adaptable underactuated two-finger gripper, successfully grasps convex and non-convex objects and supports SCALER.
\end{abstract}

\section{INTRODUCTION}
Climbing robots have promising potential for applications such as field inspections, exploration, and scientific research, and free-climbing \cite{free_climb} presents a unique challenge to robot system design. 
In this work, we focus on legged climbing robots (e.g., \cite{lemur3}) because limbed systems have high versatility and maneuverability \cite{anymal} compared to wheeled robots.
The real-world walls and cliffs are non-continuous, and climbers may have to traverse through not only a vertical wall but also the ground, overhang, or even a ceiling, i.e., in bridge inspections.

However, limbed robots are inefficient compared to wheeled vehicles since all leg joints have to be actuated and continuously support the robot's weight and load. Climbing demands more power to increase the potential energy and counter moments due to gravity to stay on the wall. If the robot needs to support payloads for practical tasks, this requirement becomes more demanding. 

\begin{figure}[t!]
    \centering
    \includegraphics[width=0.45\textwidth, trim={7cm 0cm 0cm 0cm},clip]{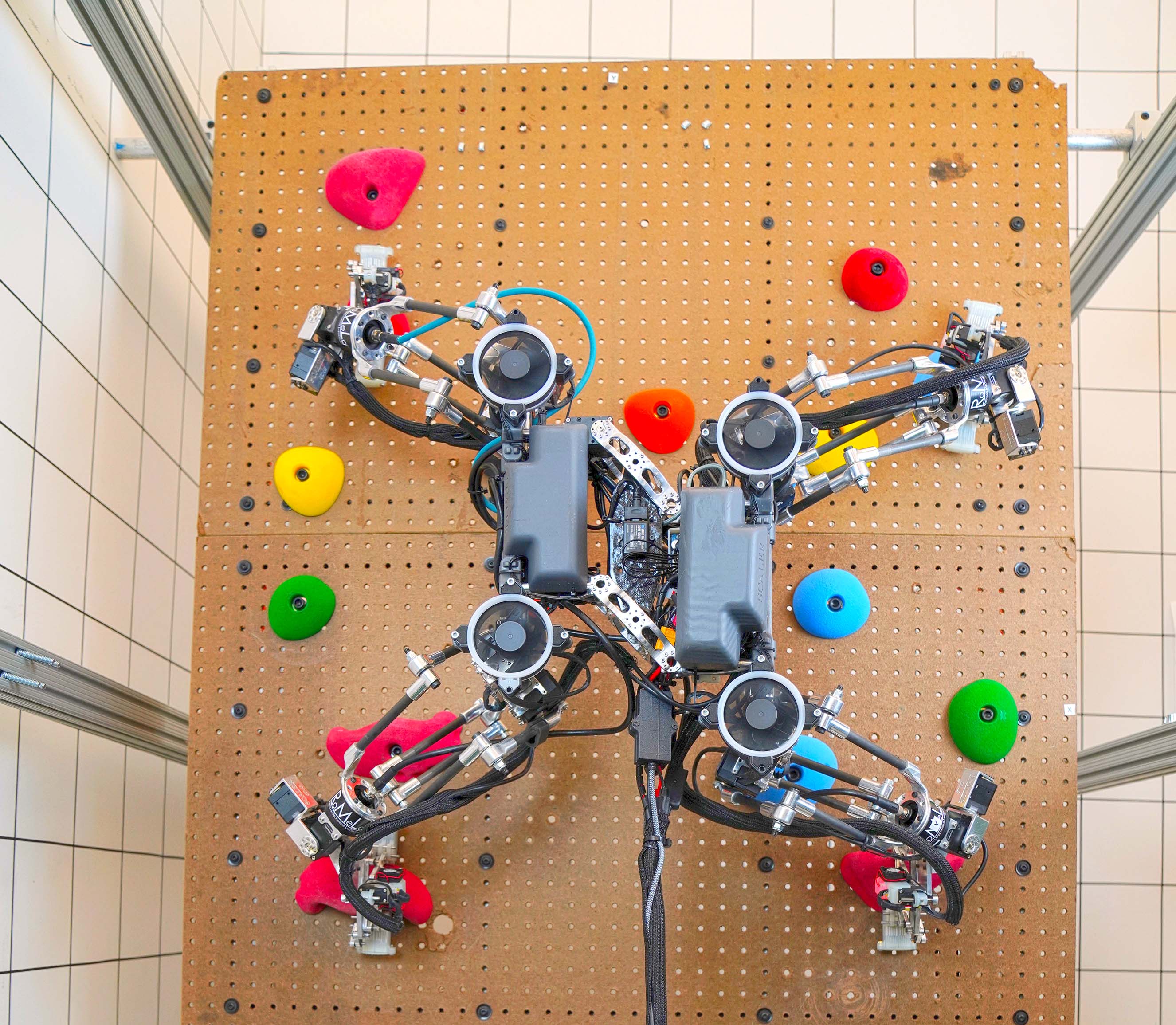}
     \caption{SCALER, a tough versatile quadruped free-climbing robot on a bouldering wall when the body is shifted. \label{fig:fig1}}
\end{figure}

In a \textit{discrete} environment seen as in \fig{fig:fig1} climbing robots have to consider foot placements and body stability \cite{ladder_climb}.
Discrete environments, such as a bouldering wall, are significantly more challenging than continuous surface or ladder climbing since robots must simultaneously conduct locomotion, manipulation, and grasping (loco-grasping) to climb successfully. Consequently, the robot platform and its gait designs must regard such unique specifications.

\begin{table*}[th!]
\centering
\begin{threeparttable}
\caption{Quadruped robot ground performance comparisons \cite{stanford_doggo}, \cite{quad_comparison}. \label{tb:comparision}}
\begin{tabular}{lcccccccc}
\hline
\multicolumn{1}{c}{Robot} & \begin{tabular}[c]{@{}c@{}}DoF \\ per Limb\end{tabular} & Weight & \begin{tabular}[c]{@{}c@{}}Normalized\\ Payload\end{tabular} & \begin{tabular}[c]{@{}c@{}}Max \\ Payload\end{tabular} & \begin{tabular}[c]{@{}c@{}}Normalized\\ Speed\end{tabular} & \begin{tabular}[c]{@{}c@{}}Ground\\ Velocity\end{tabular} & \begin{tabular}[c]{@{}c@{}}Normalized\\ Workload\end{tabular} \\ \hline
SCALER (Walking) & 3 & 6.3kg & 2.33 & 14.7kg & 1.87/s & 0.56m/s & 3.88 \\
ANYmal & 3 & 30kg & 0.33 & 10kg & 1.0/s & 0.8m/s & 0.33 \\
Stanford Doggo & 2 & 4.8kg & N/A & N/A & 2.14/s & 0.9m/s & 3.2 \\
Titan XIII & 3 & 5.65kg & 0.89 & 5kg & 4.29/s & 0.9m/s & 3.79 \\
SPOT & 3 & 30kg & 0.47 & 14kg & 1.45/s & 1.6m/s & 0.68 \\
Mini Cheetah & 3 & 9kg & *1 & *9kg & 6.62/s & 2.45m/s & N/A \\ \hline
\end{tabular}
\begin{tablenotes}
      \small
      \item *The value is theoretical based on the continuous force per legs. Normalized Velocity: body length over velocity. Normalized Payload: body weight over payload. Normalized Work Capacity: normalized speed $\times$ normalized payload, which is a mechanical efficiency metric.
    \end{tablenotes}
\end{threeparttable}
\end{table*}

\begin{table*}[th!]
\centering
\begin{threeparttable}
\caption{Recent high DoF legged climbing robot comparison. \label{tb:climbing}}
\begin{tabular}{lccccccc}
\hline
\multicolumn{1}{c}{Robot} & \begin{tabular}[c]{@{}c@{}}DoF\\ per Leg\end{tabular} & \begin{tabular}[c]{@{}c@{}}Climbing\\ Gravity\end{tabular} & \begin{tabular}[c]{@{}c@{}}Climbing\\ Velocity\end{tabular} & \begin{tabular}[c]{@{}c@{}}Normalized\\ Velocity\end{tabular} & \begin{tabular}[c]{@{}c@{}}Climbing\\ Payload\end{tabular} & End-Effector & Environment \\ \hline
\begin{tabular}[c]{@{}l@{}}SCALER\\Climbing\end{tabular} & 6 & \begin{tabular}[c]{@{}c@{}}Earth, G\\ 9.81m/s\textasciicircum{}2\end{tabular} & 0.35m/min & 1.0/min & \begin{tabular}[c]{@{}c@{}} 3.4kg\\At 0.16m/min \end{tabular} & \begin{tabular}[c]{@{}c@{}}Two finger w/ spine\\ (GOAT Gripper)\end{tabular} & \begin{tabular}[c]{@{}c@{}}Discrete, Overhang,\\ Ceiling\end{tabular} \\
\rowcolor[HTML]{EFEFEF} 
\begin{tabular}[c]{@{}l@{}}LEMUR $3$ \cite{lemur3}\\ by JPL\end{tabular} & 7 & \begin{tabular}[c]{@{}c@{}}Mars, Moon, Zero\\ 0.38G, 0.17G, 0G\end{tabular} & 0.0027m/min & *0.0067/min & \cellcolor[HTML]{EFEFEF} & \begin{tabular}[c]{@{}c@{}}Radial Velcro-like\\ Spine Gripper\end{tabular} & Continuous, Rough \\
\begin{tabular}[c]{@{}l@{}}HubRobo\cite{hubrobo}\\ by Tohoku U\end{tabular} & \begin{tabular}[c]{@{}c@{}}3\\+3 passive\end{tabular} & \begin{tabular}[c]{@{}c@{}}Mars\\ 0.38G\end{tabular} & 0.17m/min & 0.57/min & \cellcolor[HTML]{EFEFEF} & \begin{tabular}[c]{@{}c@{}}Radial Passive \\ Spine Gripper\end{tabular} & Discrete \\
\rowcolor[HTML]{EFEFEF} 
\begin{tabular}[c]{@{}l@{}}Slalom\cite{slalom}\\ by NUAA\end{tabular} & \begin{tabular}[c]{@{}c@{}}4\\+ ball joint\end{tabular} & \begin{tabular}[c]{@{}c@{}}Slope: 30$\degree$\\ $\sim$0.5G\end{tabular} & 4.2m/min & 12/min & \multirow{-5}{*}{\cellcolor[HTML]{EFEFEF}\begin{tabular}[c]{@{}c@{}}N/A \\ Not Under\\ Earth's gravity\\\end{tabular}} & \begin{tabular}[c]{@{}c@{}}Dry Adhesive\\ EPDM Rubber\end{tabular} & \begin{tabular}[c]{@{}c@{}} Continuous \\ Flat Solid/Soft \end{tabular}  \\ 

\begin{tabular}[c]{@{}l@{}}RiSE \cite{rise_bd}\\by IIT\end{tabular} & \begin{tabular}[c]{@{}c@{}}2\\4-bar link\end{tabular} & \begin{tabular}[c]{@{}c@{}}Earth, G\\ 9.81m/s\textasciicircum{}2\end{tabular} & 15.0m/min & 40.0/min & \begin{tabular}[c]{@{}c@{}} 1.5kg\end{tabular} & \begin{tabular}[c]{@{}c@{}}Spine array\end{tabular} & \begin{tabular}[c]{@{}c@{}}Continuous\end{tabular} \\
\rowcolor[HTML]{EFEFEF} 
\begin{tabular}[c]{@{}l@{}}Bobcat \cite{bobcat} \\by FSU\end{tabular} & \begin{tabular}[c]{@{}c@{}}2\\5-bar link\end{tabular} & \begin{tabular}[c]{@{}c@{}}Earth, G\\on strap\end{tabular} & 10.5m/min & 22.8/min & \begin{tabular}[c]{@{}c@{}} N/A \end{tabular} & \begin{tabular}[c]{@{}c@{}}Spine array\end{tabular} & \begin{tabular}[c]{@{}c@{}}Continuous\\Wire mesh\end{tabular}\\
\hline

\end{tabular}
\begin{tablenotes}
      \small
      \item Climbing Gravity: simulated gravity in their robot climbing tests. *LEMUR $3$ body length was approximated as the half of the track length in \cite{lemur3}. The grippers in SCALER, LEMUR 3, and HubRpbot consist of actuated DoF, which is not included in the leg DoF. 
    \end{tablenotes}
\end{threeparttable}
\end{table*}

In this paper, we propose a quadruped free-climbing robot, SCALER: Spine-enhanced Climbing Autonomous Legged Exploration Robot in Fig. \ref{fig:fig1}. SCALER is one of the first high-degrees of freedom four-limbed loco-grasping platform that can free-climb under Earth's gravity and run on the ground. The SCALER body linkage mechanism and the $\lq$spine$\lq$ fingertip GOAT gripper have allowed SCALER to climb in a discrete environment, on vertical or inverted walls and with significant payloads. SCALER performance on the ground and climbing is compared in Table \ref{tb:comparision} and \ref{tb:climbing}, respectively. SCALER is one of the most mechanically efficient quadruped based on normalized workload capacity. 
Our contributions are summarized as follows:
\begin{enumerate}
    \item We develop SCALER: a versatile quadruped limbed climber that is capable of traveling on the ground, vertical walls, overhangs, and ceilings \textit{under the Earth's gravity}.
    \item We propose a new climbing gait called SKATE gait that increases the maximum payload for climbing by utilizing the SCALER body mechanism.
    \item We apply an improved mechanically adaptive GOAT gripper with spine contact tips.
    \item We validate the SCALER robotic climber platform in hardware experiments.
\end{enumerate}

\section{Related Works}
In this section, we first go over state-of-art climbing robots. Then, we discuss previous works of climbing robot design, especially focusing on body linkage, legs, and grippers because they are key technologies for free-climbing.
\subsection{Climbing Robots}

Both wheeled \cite{wheel_climbing} and linkage mechanism-based legged climbing robots \cite{rise_bd} have successfully demonstrated climbing capabilities with spine-enhanced contacts for rock and concrete or with dry adhesive for clean flat surfaces and windows \cite{gekko_robot}. Soft robotics inspired by inchworm \cite{inchworm} or octopus \cite{octopus} have replicated unique locomotion and grasping on a small scale. A more practical climber for suspension cables has been proposed that is intended for visual inspection rather than solely climbing \cite{suspension_bridge}. Although these robots could climb vertically, their locomotion was either fixed or limited and unsuitable for carrying payloads and performing tasks.

LEMUR $3$ is a high degree of freedom (DoF) quadruped limbed robot for future space exploration missions \cite{lemur3}. HubRobo  \cite{hubrobo} has reduced the hold grasping problem by employing a spine gripper that can passively grip a hold under Mars gravity. 
However, neither can demonstrate free-climbing under the Earth's gravity.


\subsection{Body linkage Mechanism}
Body mechanisms based on animal spines have been analyzed and validated for ground and climbing robots. Cheetah spine motions are realized to improve the efficiency of the locomotion by adding compliance in the body \cite{cheetah}. Traversability in the steering is tested in  \cite{spine_horizontal} with a body rotational DoF. Inchworm gait that extends and bends body sequentially is mimicked to allow a small soft robot to climb up on a pole \cite{inchworm}.
The gecko in-plane bending movement is replicated in Slalom \cite{slalom}, which has reduced energy consumption in climbing by half compared to the rigid body on $30\degree$ slope, equivalent to $0.5$ Earth's gravity ($0.5$ G). 

\subsection{Leg Design}
The current state-of-the-art quadruped robot employs quasi-direct drive torque actuators, as seen in the MIT Cheetah \cite{cheetah}, or series elastic actuators, as seen in ANYmal \cite{anymal}, that minimizes the leg inertia to realize rapid and dynamic motions. Though the near direct drive is proven to be successful on the ground, they are not optimal in climbing which is continuously power-intensive for the duration of the operation. 
LEMUR $3$ is specialized for climbing under reduced gravity environments, which sacrifices locomotion speed and dynamic motion capability in exchange for rich continuous and stall torque provided by non-backdrivable $1200$ gear ratio servo motors \cite{lemur3}. HubRobo has demonstrated bouldering free-climbing under $0.38$ G using relatively high gear ratio dc servo motors in serial  \cite{hubrobo}. Parallel five-bar linkage mechanism-based legs used in Minitaur \cite{five_bar_leg} and Stanford Doggo \cite{stanford_doggo} are both competent in dynamic motion and have high output force because two actuators' output power is linked to one. These mechanisms have minimal inertia since the heavy actuator components are in the body. Bobcat \cite{bobcat} optimized the Minitaur link design specifically for climbing, and it has demonstrated dynamic climbing on a wire mesh. On the other hand, linkage-mechanisms increase complexity, and a singularity exists inside the practical workspace.
When one robot needs to climb, run, manipulate, and grasp objects, it is crucial to simultaneously balance power density, speed, and mechanical efficiency.

\subsection{Climbing Grippers and Spine Tips}
Grippers for climbing can be categorized by if there is explicit grasping, such as with fingers or not. Magnetic \cite{magnetic_gripper} or suction-based \cite{suction_gripper} end effectors, dry adhesive toes such as a gecko gripper \cite{gekko_robot}, and EPDM Rubber \cite{slalom} are other viable options for climbing robots. LEMUR 2B \cite{lemur3} has demonstrated a bouldering wall climbing by hocking a high-friction rubber wrapped end-effectors. The pure frictional force is sufficient when a robot is climbing between two walls \cite{multi-surface}. These implicit types can reduce the climbing loco-grasping problem down to locomotion as long as they can stick to a wall. For rough concrete, non-magnetic surfaces, or loose cloth, spine-enhanced feet are desirable, such as in Spinybot II \cite{spinybot}, CLASH \cite{clash} or in two wall climbing \cite{risk_aware}. Spines, or needles, can get inserted or `spine' into microcavities of rocky surfaces \cite{hubrobo}. 
Explicit gripping in climbing is explored in LEMUR $3$ and HubRobo. They consist of radially aligned micro-spines, supporting the robot weights under reduced gravity.
A more explicit object grasping-based climbing gripper, called SpinyHand \cite{spiny_hand}, has four underactuated tendon-driven fingers with an ability to switch between crimp and pinch grasping pose and has exhibited promising results for a human scale climbing robot. In a discrete climbing environment, the allowable margin of the error is significantly constrained due to the loco-grasping problem. Therefore passive and mechanical solutions against stochasticity in operation are valuable.

\section{SCALER Hardware\label{sec:hardware}}
 \begin{figure}[h!]
    \centering
    \includegraphics[width=0.49\textwidth, trim={0cm 0cm 1cm 0cm},clip]{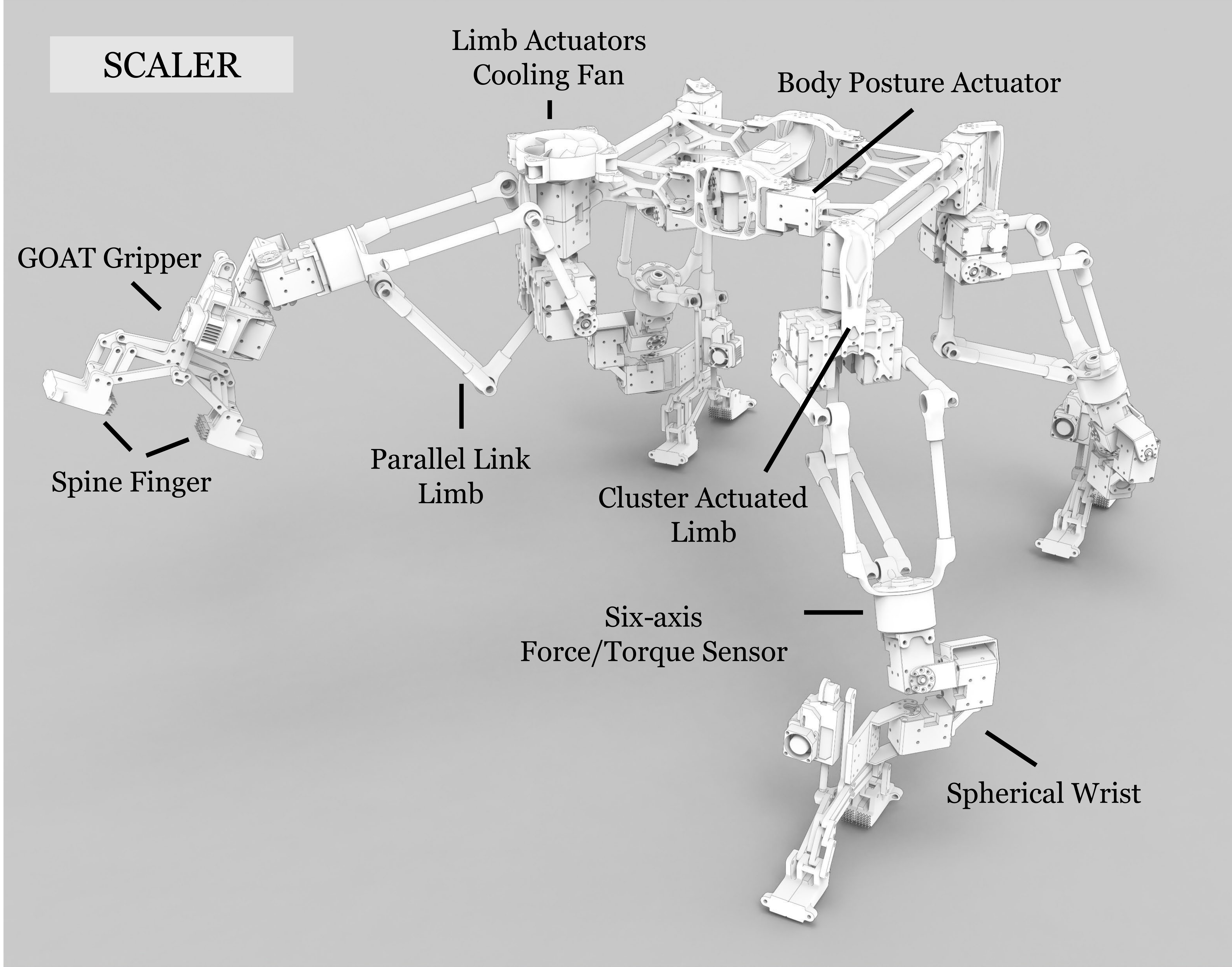}
     \caption{SCALER isometric view. SCALER is a four-limbed climbing robot with six-DoF legs and underactuated GOAT grippers with one actuator each. The body posture actuator adds 1-DoF translation motion in the body.  \label{fig:SCALER}}
\end{figure}

SCALER in Fig. \ref{fig:fig1} and \ref{fig:SCALER} is designed to achieve tightly coupled loco-grasping capabilities and to meet actuation power density and robust operation requirements in climbing tasks while maintaining dynamic and fast mobility on the ground. Those specifications are accomplished with three key technologies: body shifting mechanism, clustered parallel link leg, and mechanically adaptable spine gripper.

\subsection{Body Mechanism}
\subsubsection{Body Posture Actuation}
The SCALER body employs a four-bar linkage mechanism driven by one actuator in Fig. \ref{fig:one_leg_body_shift_gait}, which improves flexibility and payload. Conventional four-legged robots involving rigid one-body and body translation DoF are rarely adapted due to limited benefits in locomotion in contrast to the complexity introduced. However, the body-shifting DoF can play a significant role where locomotion and grasping coincide since it grants the potential to shift the robot workspace on demand. Translation body motion can better scale to increase stride length than extending leg lengths since the limb workspace is spherical. 
In climbing tasks, a multi-body system can enhance the payload instead of lifting the entire rigid body at once by utilizing the body posture actuator's power and halving the load to raise at a time. 
Another link at the body center in Fig. \ref{fig:one_leg_body_shift_gait} is stationary with respect to the SCALER body frame. The stationary body link houses an IMU and battery compartments which help concentrate mass to the center and reduce effects on dynamics when shifting the body.

\subsubsection{SKATE Gait\label{Body}}
\begin{figure}[h!]
    \centering
    \includegraphics[width=0.49\textwidth, trim={0cm 0cm 0cm 0cm},clip]{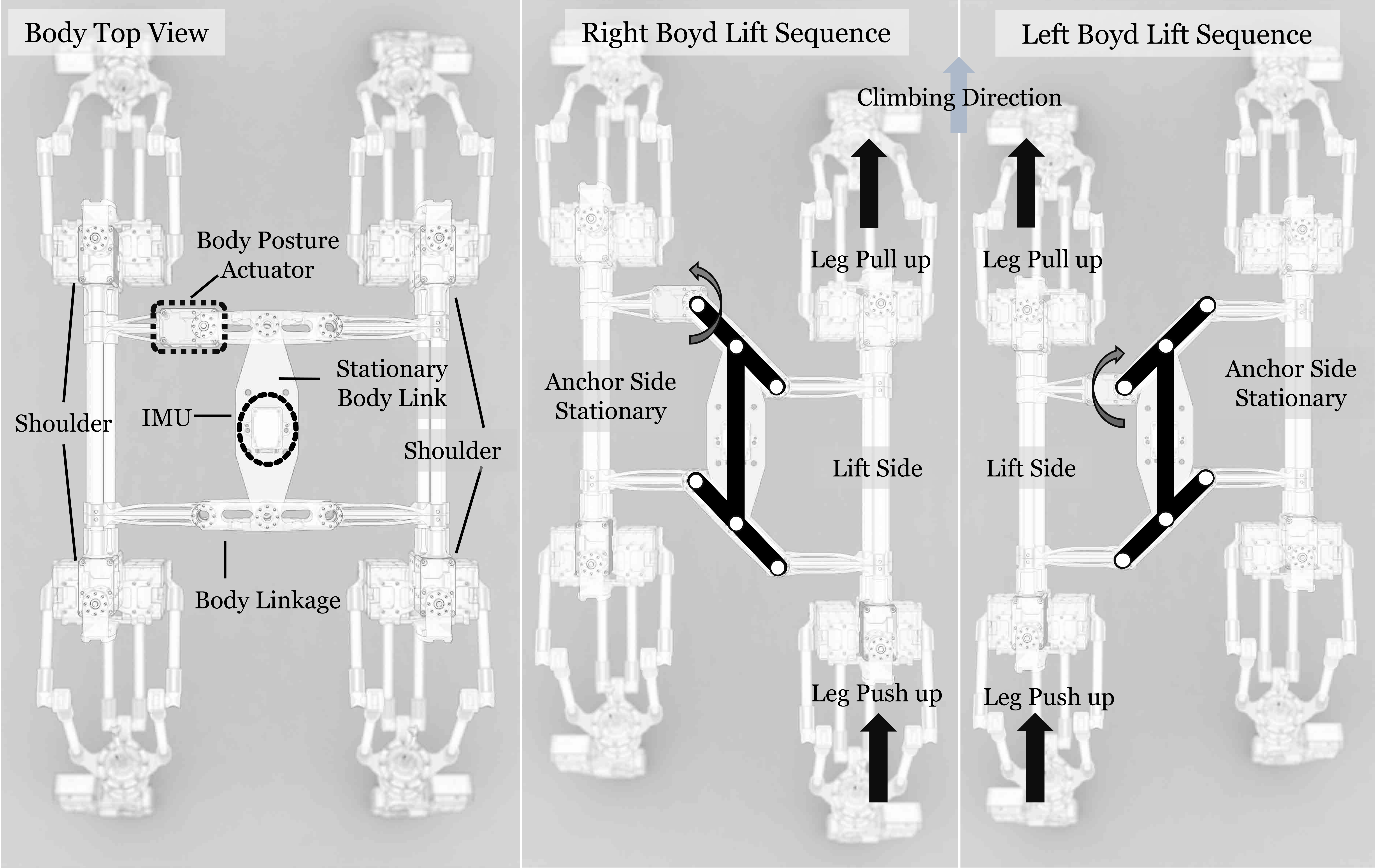}
     \caption{SKATE gait. Black bars and white dots represent kinematic link design and joint placements. By alternately lifting the half of the body (switching kinematics between right and left configuration in the figure), SCALER can utilize the body posture actuator to improve the payload. \label{fig:one_leg_body_shift_gait}}
\end{figure}

SCALER's body shifting mechanism allows a unique and progressive gait, which we call: Shifting Kinematics Adaptive Torso Extension (SKATE) gait:
\begin{enumerate}
    \item Move forward the front limb in the lift side (the left state in Fig. \ref{fig:one_leg_body_shift_gait})
    \item Advance the lift side of the body by rotating the body posture actuator while pulling and pushing it with the lift side front and back limbs, respectively
    \item Move forward the back limb in the lift side
    \item Swap the lift side and the anchor side (the right state in Fig. \ref{fig:one_leg_body_shift_gait}) and repeat the sequence $1-3$
\end{enumerate}

The leg actuators can withstand higher torque at no motion than in continuous motion. Furthermore, moving sets of legs are aided by force from the body posture actuator. SKATE gait can adequately improve the payload on climbing robots, and the maximum load is constrained by the force of the stationary legs and the body shifting mechanism. 
For SCALER, the SKATE gait provides an additional $30$ N thrust force in a climbing direction from the body posture actuator.

\subsection{The Cluster Actuated Parallel-serial Link Limb}
\begin{figure}[h!]
    \centering
    \includegraphics[width=0.49\textwidth ,clip]{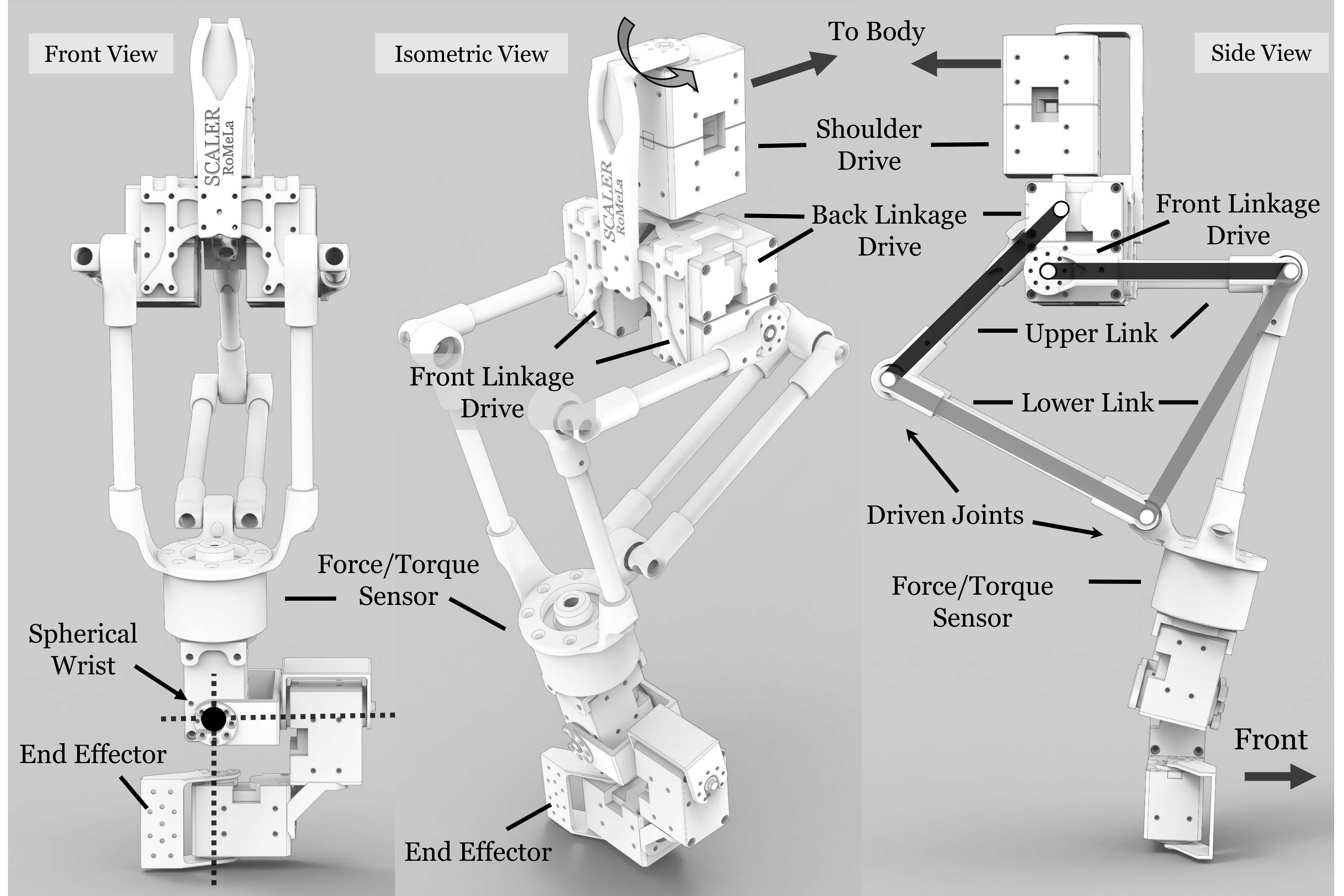}
     \caption{SCALER 6-DoF limb. The right figure shows the five-bar linkage kinematic design. The upper and lower links are in black and grey, respectively. The shoulder and five-bar linkages are driven by two motors each. The six-axis F/T sensor is installed at the end of the five-bar link, and the spherical wrist is on the sensor measurement face. 
     \label{fig:leg_design}}
\end{figure}

SCALER limb involves a 2-DoF five-bar linkage combined with a shoulder joint and the spherical serial wrist to attain 6-DoF per leg in Fig. \ref{fig:leg_design}.  
SCALER employs medium to high gear ratio actuators, Dynamixel XM430-350, that are clustered to achieve both functional speed and payload in climbing. Five-bar parallel link and shoulder joints are driven by a pair of the same actuator in Fig. \ref{fig:leg_design}, and thus six motors drive three shoulder-leg joints.

Climbing requires significant power, and actuator heat dissipation is critical. SCALER has to support and lift the system mass and withstand moments due to gravity that would otherwise pull SCALER off the wall. A very high gear ratio, as seen in LEMUR $3$ \cite{lemur3} sacrifices motion speed significantly, whereas larger and more powerful motors can still overheat. The five-bar parallel link mechanism combines two joints, each comprised of two actuators, to realize 2-DoF motions aligned in the direction of climbing.
The symmetric five-bar design used in SCALER is mechanically superior in its proprioceptive sensitivity, force production, and thermal cost of force among serial and two different five-bar linkage leg designs in \cite{five_bar_leg}. Although this choice of leg mechanism is ideal for power-intense climbing operation, the parallel mechanism suffers from singularities within the workspace, such as where both front and back elbow-driven joints are at the same axis. The SCALER's back linkage in Fig. \ref{fig:leg_design} is marginally shorter to shift such a condition mechanically out of regular operation workspace (i.e. inside the body).
In addition to the minimal thermal cost of force in the limb design, the dual motors on each joint serve to spread the heat source, keeping the motor temperature optimal and avoiding thermal throttling or damage. Furthermore, this design doubles the torque without changing the rotational velocity. The six shoulder-leg actuators are cooled by the fan above in Fig. \ref{fig:SCALER}. 

The current state-of-art four-legged robots focus on reducing the leg weights by locating actuators in the body to decrease the leg inertia and make the robot dynamics closer to the single mass model \cite{cheetah}, \cite{stanford_doggo}. However, each limb doubles as both a supporting leg and manipulation arm in climbing, and the leg design has to be optimized for unique hybrid dynamics motions. SCALER needs to raise end effectors and the body alternately as it climbs. Accumulating limb mass to the body promotes manipulation or swing leg dynamics, but the rest of the limbs have to suffer more weight. Consequently, to balance these contradictions, SCALER limb weight has been distributed on both ends: the body and wrist. The intermediate links are structured with carbon fiber tubes. The leg inertia becomes optimal for walking once grippers and wrists are removed.

\subsection{Modularity}
SCALER consists of four limbs, 6-DoF each, 1-DoF in the body, and in total, 25-DoF, which are driven by overall 37 DC servo motors, equipped with two-finger underactuated 2-DoF grippers driven by one DC linear actuator each. SCALER's legs and body are modular designs and easily replaced for replacements or different configurations such as walking or manipulator modes. SCALER's walking configuration has 3-DoF per leg by replacing the spherical wrist with a semi-spherical foot cover to protect the force/torque (F/T) sensor, which is pictured in Fig. \ref{fig:trot_weight}. 
The walking format benefits from the minimal inertia design of SCALER's linkage limb since most of the leg weight is from the wrists and GOAT grippers. 

\section{Gripper}
SCALER is equipped with GOAT grippers \cite{GOAT} in Fig. \ref{fig:GOAT}, which is a mechanically adaptable whippletree-based underactuated rigid two-finger gripper. Manipulation and grasping must be conducted precisely under various uncertainties to successfully climb up in discrete environments. The GOAT gripper can mechanically compensate for end-effector position errors with one passive DoF.

\subsection{GOAT Mechanism Gripper}
\begin{figure}[h!]
    \centering
    \includegraphics[width=0.4\textwidth, trim={0cm 0cm 0cm 0cm},clip]{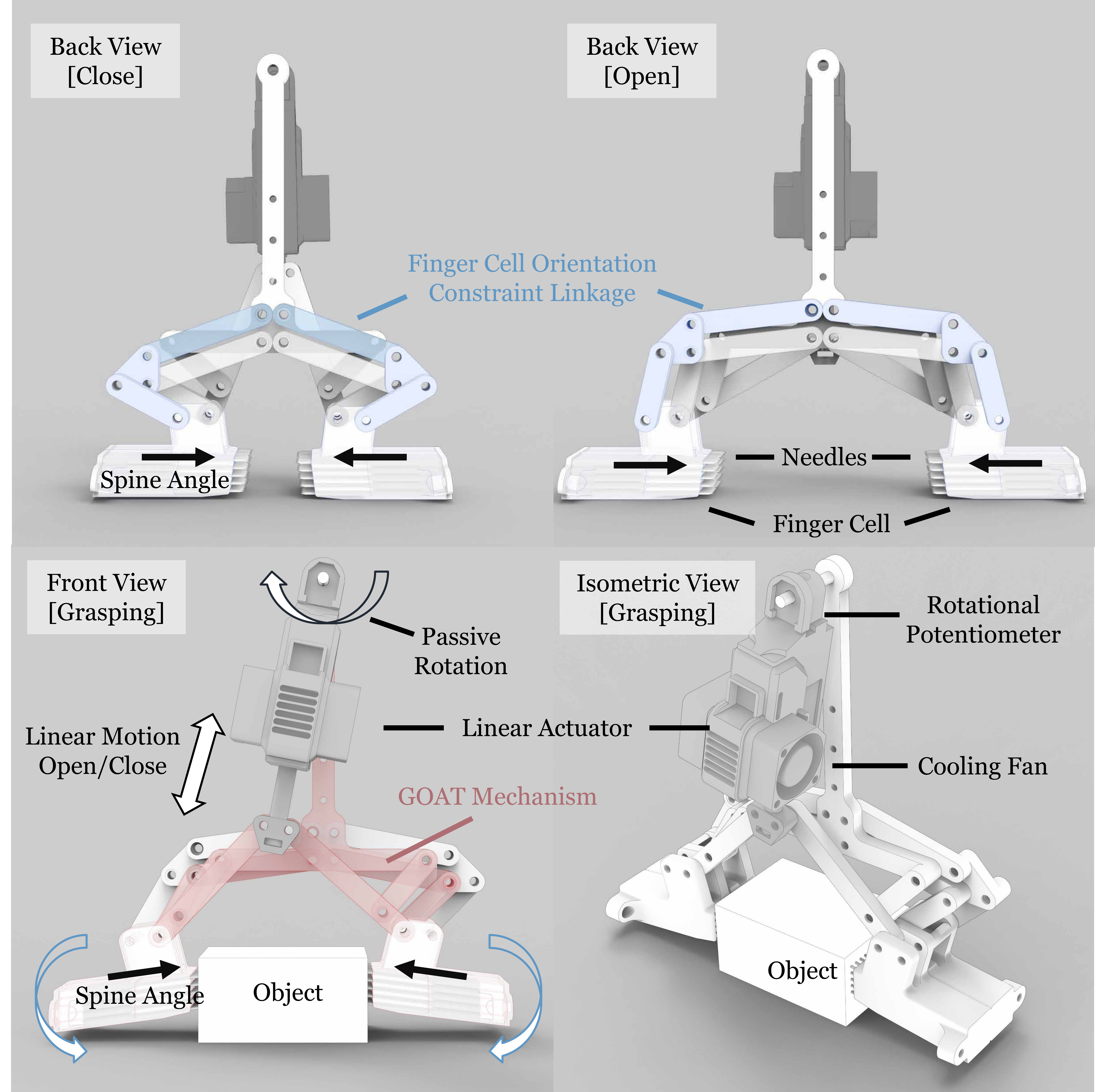}
     \caption{GOAT Gripper with four-bar linkages constraining finger cell orientation. The spine needle angle of attack is optimal when grasping a flat object, as shown in the bottom left figure. Details about GOAT linkage mechanism are in \cite{GOAT}.  \label{fig:GOAT}}
\end{figure}

Our GOAT gripper utilizes the GOAT mechanism based on the whippletree load balancing linkage system \cite{GOAT}. The end effector positions may be off due to gravity effects or uncertainty in the system. Conventional two-finger grippers grasp an object at their central axis since they only have 1-DoF, open or closed. In climbing, gravitational force often sags the grippers downward. If a jaw gripper is evenly closed, such offset may translate into undesired internal elastic energy if the grasping is successful, to begin with. The GOAT gripper, instead, has 2-DoF, which can passively translate horizontally in the gripper plane. Thus the GOAT mechanism can mechanically adapt to the off-center object and grasp it \textit{as is}. 
GOAT gripper linkage lengths were optimized in \cite{GOAT} for the set of bouldering holds in our environment and force requirements to support SCALER weights. 
The GOAT gripper is upgraded with additional linkages constraining finger cell orientation and more powerful linear actuators.
SCALER's GOAT gripper finger cell orientations are constrained to be parallel to each other with external four-bar linkages as shown in \fig{fig:GOAT}, which fix spine needle angle of attacks at optimal, regardless of GOAT kinematics configurations.

\subsection{Spring-loaded Spine Cell Fingers}
Our spring-loaded spine cell design is based on \cite{spine_cell}. Each cell consists of fifty $\diameter 0.93$ spines, each loaded by a $5$ mN/mm spring. The cell surface is slanted so that the spines approach at an optimal angle. The slight angular compliance improves spine angle of attack when grasping, whereas it helps to detach spines from microcavities when releasing.

\subsection{Grasping Force Controllers}
Grasping force, or normal force at each fingertip, is valuable to improve climbing stability and reduce the duty ratio of the GOAT drive motor, which has to provide linear force to grasp constantly. 
The GOAT gripper can be force-controlled with either the current-force or stiffness-based control. SCALER's GOAT gripper is operated using a MightyZap linear actuator by IR Robot, which has current-based force control. Input force to the GOAT mechanism can be converted into output fingertip force using static equilibrium since the linear actuator is relatively slow. Stiffness model data is collected using a load cell while grasping different object sizes since the input-output force is nonlinear. 
GOAT gripper maximum withstanding force has been linearly modeled against object surface slope using Gaussian process regression in \cite{GOAT}, representing force upper bound the gripper can hold instead of a traditional friction cone that is nonlinear due to spine tip effects \cite{spine_cell}.


\section{SCALER Software \label{sec:software}}
\begin{figure}[h!]
    \centering
    \includegraphics[width=0.4\textwidth ,clip]{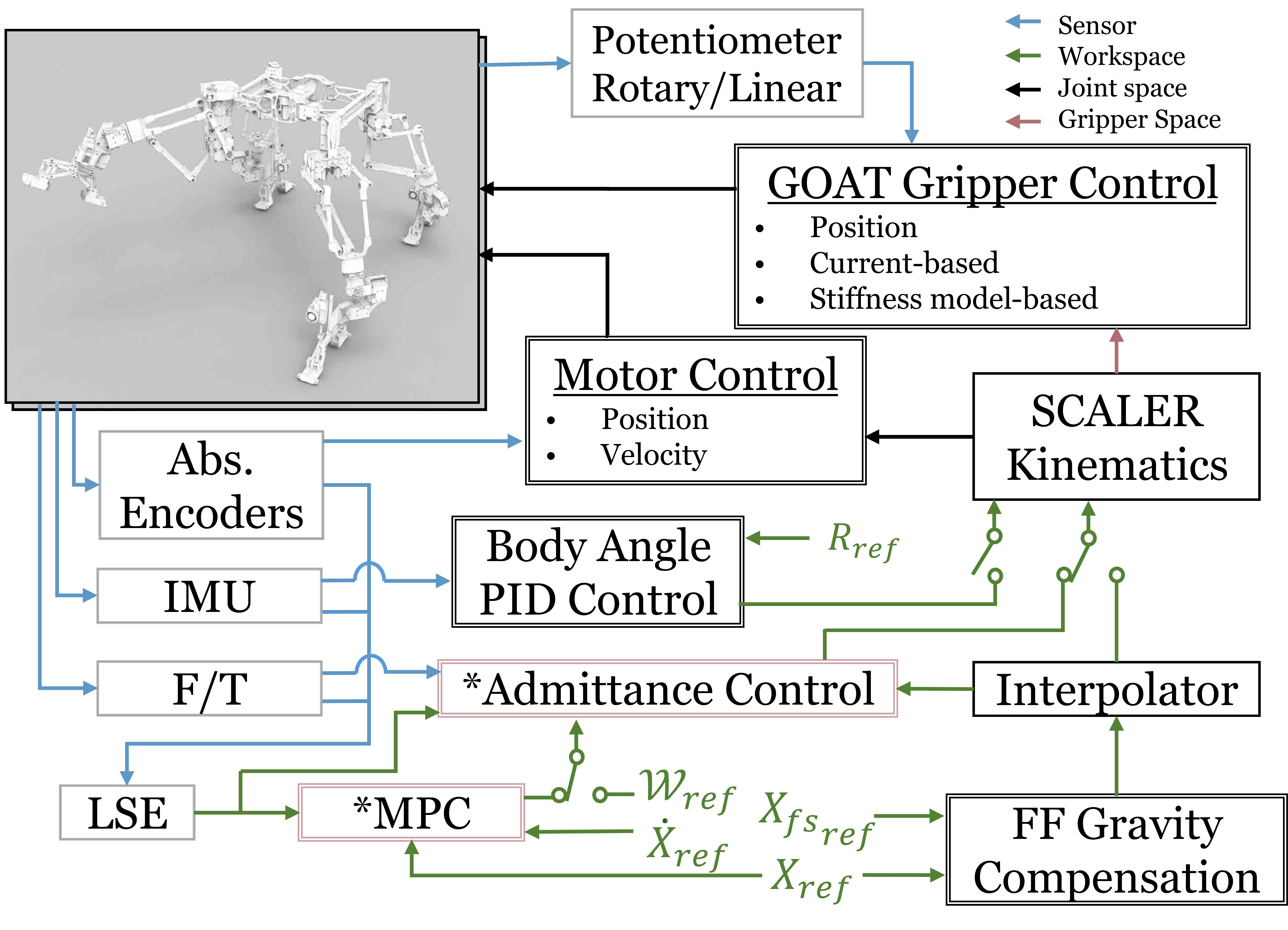}
     \caption{SCALER software architecture. $X_{ref}$ and $\dot{X}_{ref}$ are the reference trajectory of the body and toe states, and their derivatives, $X_{fs_{ref}}$ is foot steps and the body states, $\mathcal{W}_{ref}$ is wrench, and $R_{ref}$ is body orientation reference. LSE \cite{estimation} is the state estimator. The force controller stream identified by * is activated when either a wrench trajectory is provided or MPC is used. The body angle controller is not necessary if MPC is running.}
\label{Fig:software}
\end{figure}

 \begin{figure*}[th!]
 \centering
    \begin{subfigure}{0.3\textwidth}
\includegraphics[width=\textwidth,trim={0cm 0cm 0cm 0cm}, clip]{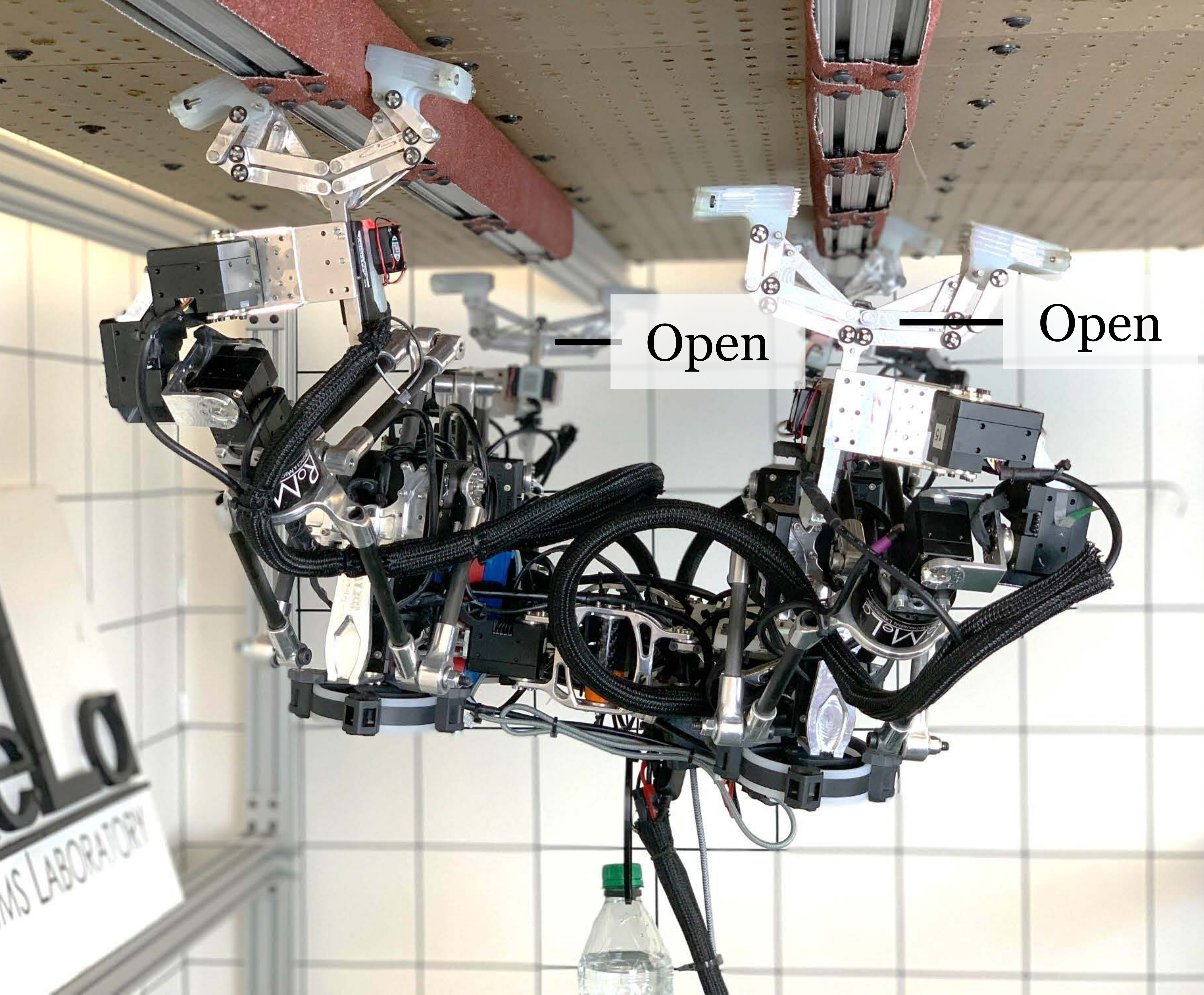}
\caption{Two finger support on the ceiling.\label{fig:ceiling_two}}
    \end{subfigure}
     \begin{subfigure}{0.3\textwidth}
         \centering
    \includegraphics[width=\textwidth,trim={0cm 0cm 0cm 0cm}, clip]{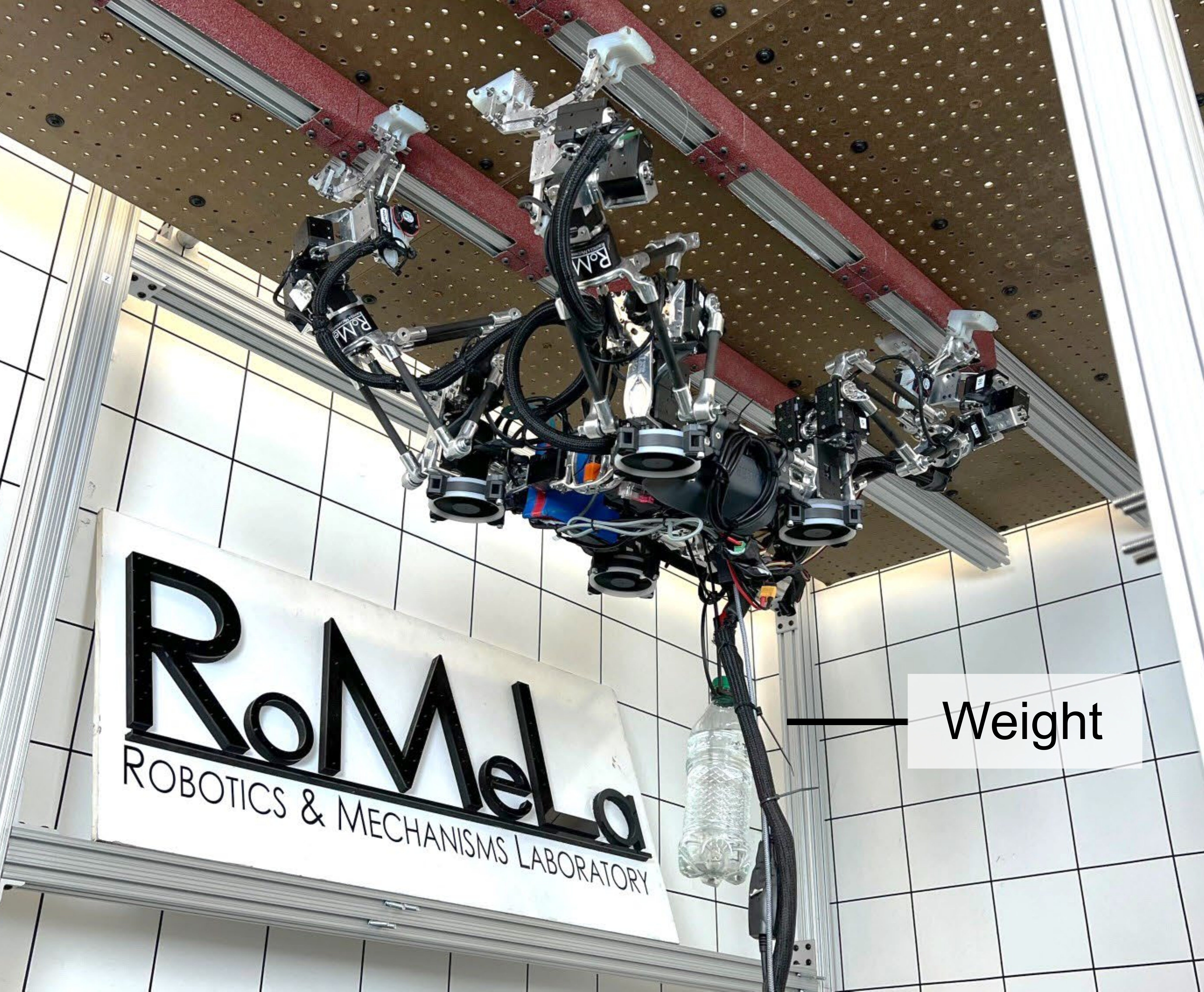}
    \caption{Ceiling SKATE gait. \label{fig:ceiling_SKATE}}
     \end{subfigure}
      \begin{subfigure}{0.3\textwidth}
    \centering
\includegraphics[width=\textwidth,trim={0cm 0cm 0cm 0cm}, clip]{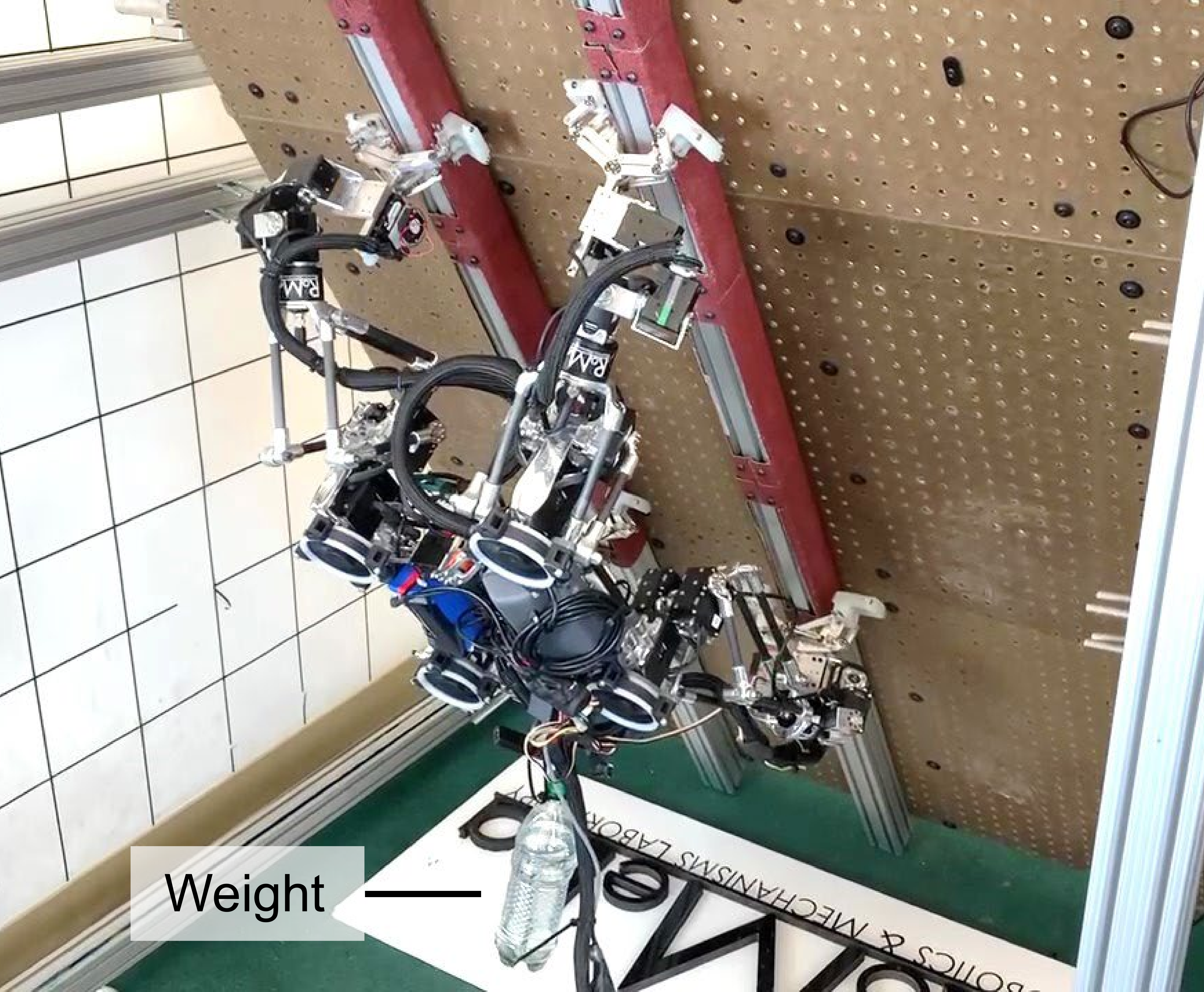}
\caption{Climbing on an overhang wall at $120\degree$.\label{fig:overhang}}
    \end{subfigure}
     \caption{SCALER under inverted environments. Weight is suspended to indicate the direction of gravity.
     }\label{fig:inverted}
\end{figure*}

SCALER is equipped with various sensors: four F/T sensors on the wrists, a body IMU sensor, Dynamixel actuator encoders, and sensors on the GOAT gripper, all of which are handled in a corresponding dedicated I/O thread.

The Dynamixel servos and GOAT gripper modules are connected over RS-485 and to the CPU via a USB bus. Each limb sustains an independent and one redundant communication-power chain to ensure robust operation in climbing that can prevent SCALER from falling due to damaged wires. Due to a large number of Dynamixel motors, the communication frequency with the encoders is capped at $150$ Hz, which is sufficient for the tasks targeted in this paper. All other sensors run at over $400$ Hz.

Joint space position control is either done on the Dynamixel actuator or on the CPU, which runs our custom position controllers that generate velocity control inputs to the actuators. Though the communication frequency limits the control frequency, joint velocity control allows SCALER to perform more aggressive motions.

Admittance control \cite{alex_admittance} is implemented as operational space force control to track a contact reaction force or wrench trajectory by extending \cite{admittance}.
Due to gear backlash, the force control struggles to track rapidly changing force profiles (e.g., a square wave of more than $10$ Hz). This control bandwidth suffices for low-speed tasks such as climbing that experience less frequent force reference changes.

Legged State Estimation (LSE) is implemented from \cite{estimation} which takes IMU, FT sensor, contact detection, and encoders along with kinematics and provides pose and velocity estimations using an Extended Kalman filter. The state estimation is used for an MPC controller based on \cite{MPC} that can be used instead of explicit wrench references.  A diagram of software architecture is shown in Fig. \ref{Fig:software}. SCALER body and foot trajectories can be generated either manually or by optimization-based planners \cite{contact_rich} outside of the current SCALER software structure in Fig. \ref{Fig:software}.

On a bouldering wall, feasible grasping regions are very sparse since the bouldering holds are the only candidates for gripping. Though a dense point cloud is a versatile way of representing 3D space, most points are unnecessary for our planning purposes. To improve the efficiency of the mapping, we constructed a sparse discrete map (SDM) that represents holds as ellipsoids and walls as planes. Hold shapes are generally non-convex, but ellipsoid fitting can conservatively estimate the sizes within $12$ \% for holds that resemble an ellipsoid in shape. We find the maximum-volume inscribed ellipsoid for each handhold which ensures that planned grasps will be feasible, regardless of the holding shape. We measure variance of $4$ mm for the ellipsoid centroid as the camera moves to view different angles of the handholds. The variance of both ellipsoid semi-axes and centroid measurements is less than the mechanical adaptability of the GOAT gripper. A RealSense D435i camera detects handholds by performing 2D image segmentation and then projecting each segmented mask into a point cloud.
We use a RealSense T265 tracking camera, which is rigidly attached to the D435i, to localize the holds in world space instead of the LSE, which can have a significant drift. 
Currently, the SDM pipeline runs at approximately $3$ Hz, and it is independent of the SCALER software framework, but external planners can utilize this simplified convex map for future works.

\section{Experiment}

\begin{table}[h!]
\caption{List of Experiments and Objectives \label{tb:experiments}}
\begin{tabular}{ccc}
\hline
Experiment                                                        & Environment                                                                                    & Objective                                                                                   \\ \hline
\begin{tabular}[c]{@{}c@{}}Bouldering\\ Sec. VI.A\end{tabular}    & \begin{tabular}[c]{@{}c@{}}Discrete, Bouldering,\\ Vertical\end{tabular}                       & \begin{tabular}[c]{@{}c@{}}Verify Climbing Velocity\\ Loco-grasping Capability\end{tabular} \\
\rowcolor[HTML]{EFEFEF} 
\begin{tabular}[c]{@{}c@{}}SKATE Gait\\ Sec. VI.B\end{tabular}    & \begin{tabular}[c]{@{}c@{}}Continuous, Bars\\ Sandpaper, Vertical\end{tabular}                 & \begin{tabular}[c]{@{}c@{}}Climbing Payload,\\ Gait Demonstration\end{tabular}              \\ \hline
\begin{tabular}[c]{@{}c@{}}Overhung\\ Sec. VI.C\end{tabular}      & \begin{tabular}[c]{@{}c@{}}Continuous, Bars,\\ Sandpaper, $125 \degree$\end{tabular} & \begin{tabular}[c]{@{}c@{}}Traversability on\\ an Inverted Wall\end{tabular}                \\
\rowcolor[HTML]{EFEFEF} 
\begin{tabular}[c]{@{}c@{}}Ceiling\\ Sec. VI.D\end{tabular}       & \begin{tabular}[c]{@{}c@{}}Continuous, Bars,\\ Sandpaper, Ceiling\end{tabular}                 & \begin{tabular}[c]{@{}c@{}}Inverted Gravity\\ Gripper Capability\end{tabular}               \\ \hline
Trot, Sec. VI.E                                                   & Ground                                                                                         & Ground Velocity,                                                                            \\
\rowcolor[HTML]{EFEFEF} 
\begin{tabular}[c]{@{}c@{}}Trot, Payload\\ Sec. VI.F\end{tabular} & \begin{tabular}[c]{@{}c@{}}Ground\end{tabular}                       & \begin{tabular}[c]{@{}c@{}}Ground Payload,\\ Velocity w/ Payload\end{tabular}               \\ \hline
\end{tabular}
\end{table}

 \begin{figure}[h!]
    \centering
    \includegraphics[width=0.49\textwidth,clip]{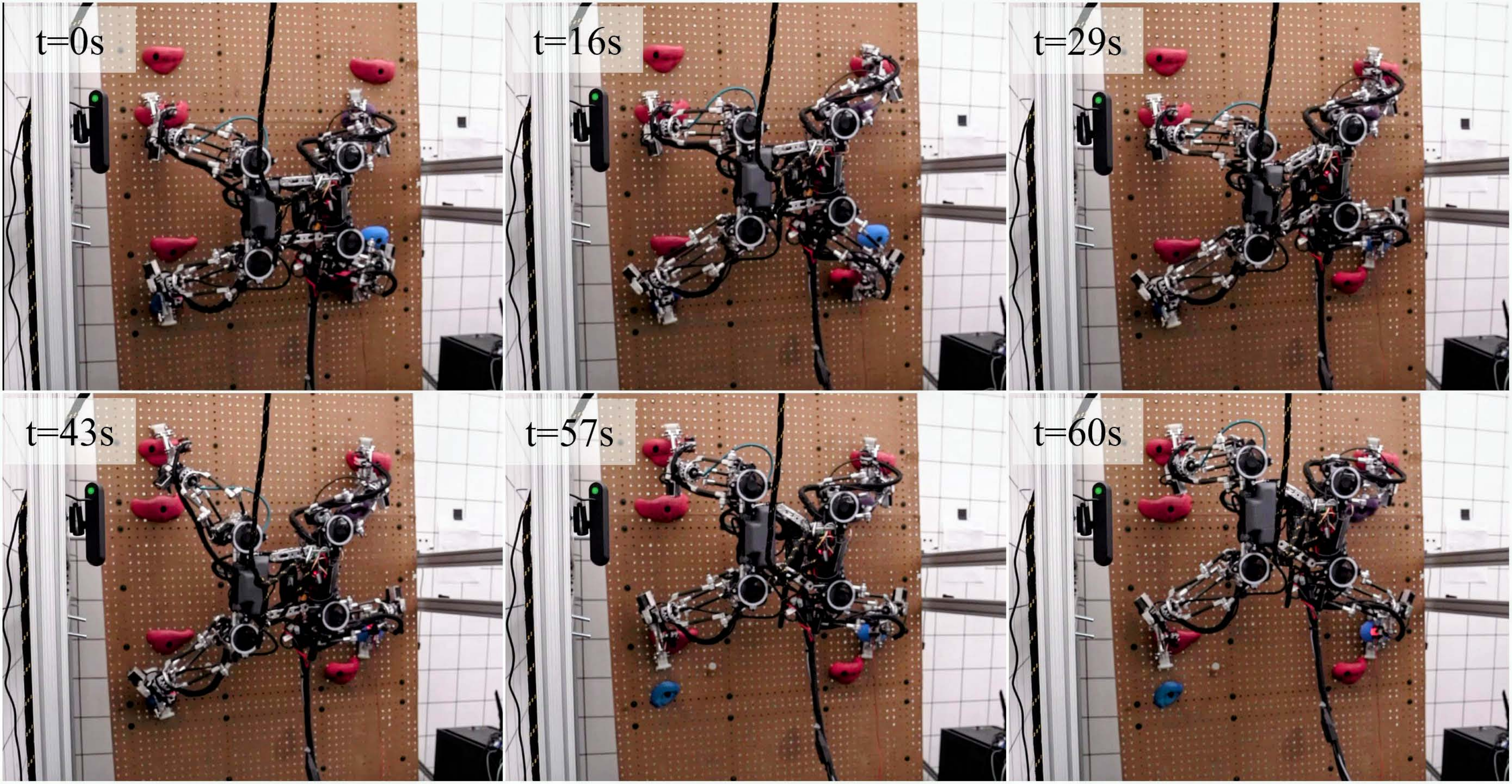}
     \caption{SCALER climbing a bouldering wall after each leg moved forward. The wall is at $90 \degree$. SCALER climbed up $0.35$ m at $0.35$ m/min or normalized speed of $1.0$ /min. The body length is slightly less than the eye to eye distance of the detachable fan on each shoulder. \label{fig:bouldering}}
\end{figure} 

The SCALER capabilities and performance were evaluated and demonstrated through discrete and continuous environments and ground tests listed in Table \ref{tb:experiments}. SCALER can climb up discrete bouldering, vertical, and overhanging walls while walking on the ground and the ceiling. For these experiments, SCALER was operated through an off-board computer and with an external power supply as a precaution. 

Table \ref{tb:comparision} compares SCALER walking configuration (3-DoF limbs) to other ground limbed quadruped robots. SCALER can reach a comparable normalized speed against ANYmal and SPOT. For payload, no other quadruped carried double the robot weight in \cite{quad_comparison}. Normalized work capacity (normalized speed $\times$ normalized payload), which serves as a metric for mechanical efficiency, exceeds that of Titan XIII, which is one of the best in \cite{quad_comparison}. The normalized work capacity is a more appropriate metric for comparing SCALER to others since it considers payload and velocity.

Table \ref{tb:climbing} lists limbed climbing robots and their system and capabilities. Other high-DoF robots in this category operate under reduced gravity. In contrast, SCALER can carry payloads under Earth G. Slalom, Rise, or Bobcat is significantly faster than the others since the environment is flat continuous terrain that does not require explicit grasping at all or the problem is reduced to locomotion. SCALER's climbing speed is mainly constrained by the gripper, which takes $8$ sec to fully open and close.

\subsection{Bouldering Vertical Free-Climbing}
Conventional polymer-made bouldering holds were installed to a $90 \degree$ wall, and SCALER was operated using a predefined manual trajectory. A feedforward controller was used in the workspace to compensate for gravitational sag down described in Section \ref{sec:software}. SCALER successfully climbed up for four steps in Fig. \ref{fig:bouldering}; each leg moved to the next bouldering hold at $0.35$ m/min or a normalized speed of $1.0$ /min. The velocity was measured based on the body travel distance.   
SCALER conducted locomotion, manipulation, and grasping simultaneously in a discrete environment.

The body posture actuator stretched the legs occasionally to shift the workspace. 
The SKATE gait can be applied to this case, although it becomes a multi-objective optimization problem between kinematics and dynamics which may require a proper planner. The SKATE gait enhances dynamics (i.e. the payload capacity), but kinematic space is a notable constraint in a discreet environment.


\subsection{Vertical Wall SKATE Climbing with Payload}
\begin{figure}[h!]
    \centering
    \includegraphics[width=0.45\textwidth,clip]{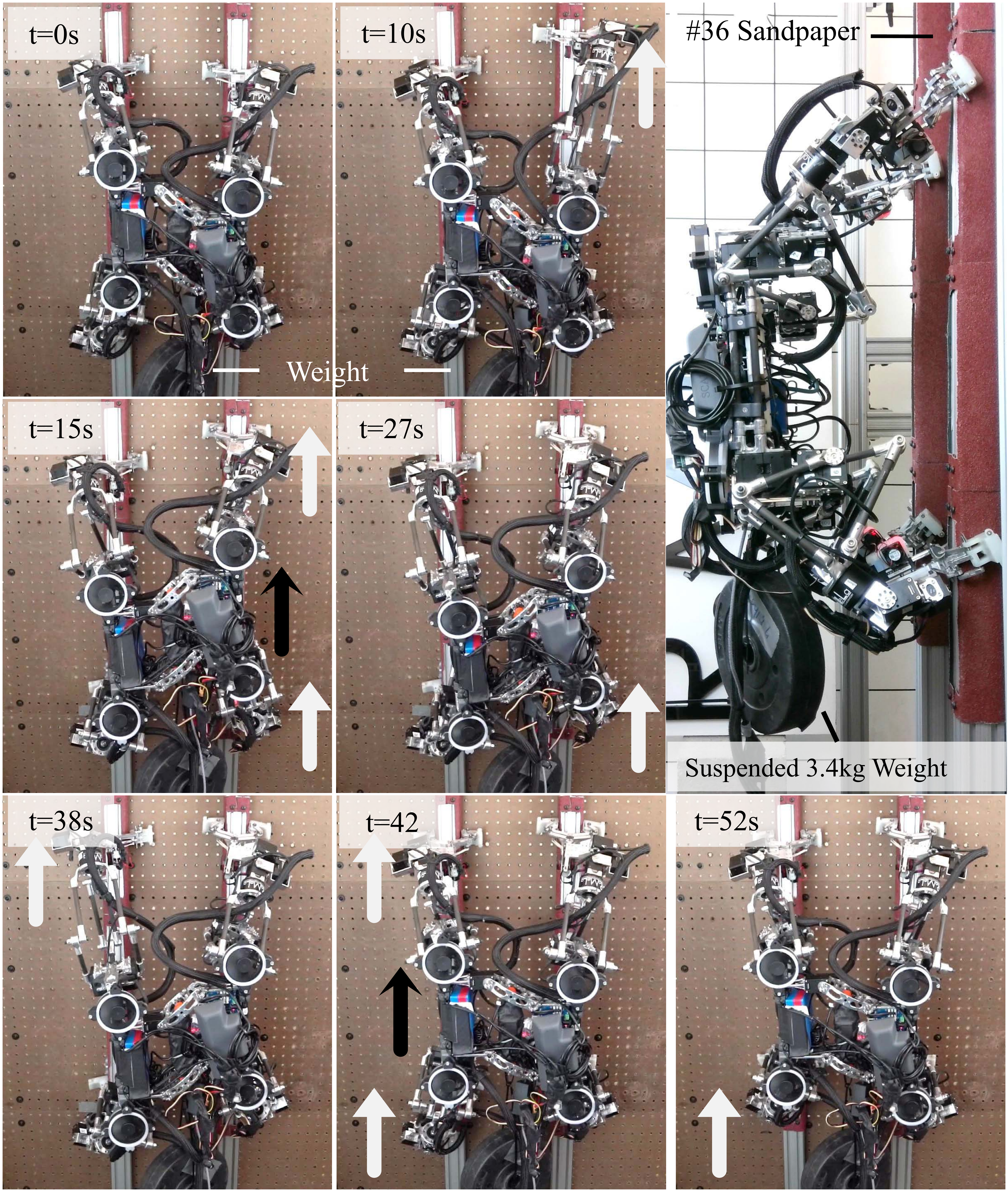}
     \caption{SKATE gait in hardware. SCALER climbing on a vertical wall carrying $3.4$ kg weight with SKATE gait. White arrows and black stealth arrows indicate leg and body motion, respectively. At $10$ s, the front-right leg moves forward. At $15$ s, The body posture actuator, the front-right and back-right legs lift the right half of the body while the left side is an anchor (stationary). At $27$ s, the back-right leg moves forward. Then, the same procedure is repeated for the left half of the body. The SKATE gait sequence lasted for $52$ s and moved $0.14$ m (i.e. $0.16$ m/s).
     \label{fig:vertical_weight}}
\end{figure}

\begin{figure}[h!]
    \centering
    \includegraphics[width=0.49\textwidth,trim={0cm 0cm 0cm 0cm},clip]{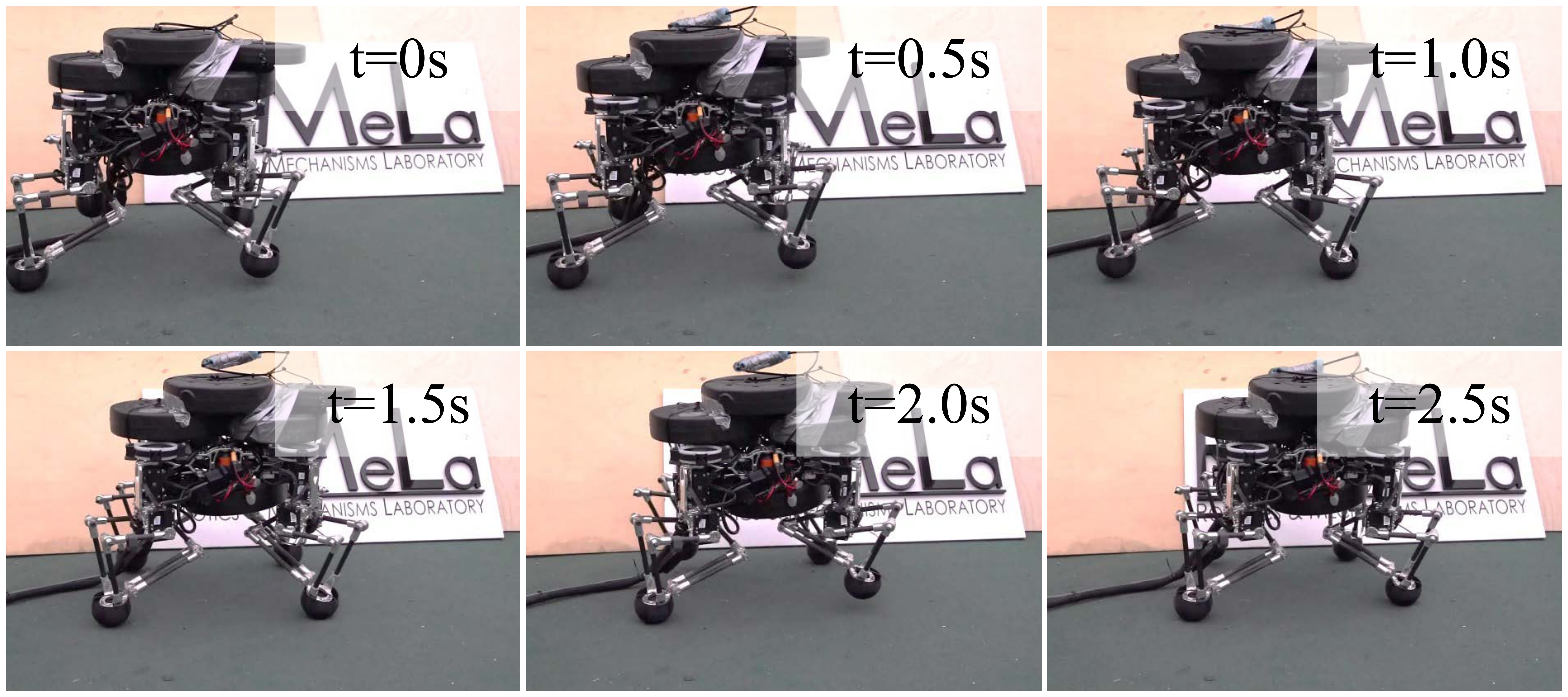}
     \caption{SCALER walking configuration (3-DoF limbs) trotting on the ground with $14.7$ kg payload (i.e., $233$ \% of its own weight). 
    \label{fig:trot_weight}}
\end{figure}

Body shifting mechanism and SKATE gait were evaluated on a vertical wall with a continuous straight two-inch rail where grasping faces were covered with $\#36$ sandpaper to mimic a rocky surface. Each body lift in SKATE gait was $0.075$ m at the body. Climbing speed was $0.16$ m/min with $3.4$ kg weight hanging on to the body center, which was $35$ \% of the SCALER weights including grippers, turning the overall SCALER weight to $13$ kg. Suspended payload is tougher than fixed weight because of the pendulum dynamics. 
The black and white arrows in Fig. \ref{fig:vertical_weight} indicate which toes and body moved compared to the previous time frame, respectively.

\subsection{Overhang Free-Climbing}
SCALER climbed on an overhanging wall at $125\degree$ tilted toward the robot in Fig. \ref{fig:overhang}. A weight of $0.5$ kg was suspended to indicate the direction of gravity. SCALER had to hang onto the wall to counter the larger moment and gravitational force that directed the robot to detach compared to the previous vertical wall.

\subsection{Ceiling Walking}
Where the robot is walking on the ceiling, GOAT gripper grasping force is critical to support the entire SCALER weight, whereas the robot experiences a similar force on the ground but upside down. SCALER was successfully grasped onto the same rail attached to the ceiling and performed body SKATE gait walking in Fig. \ref{fig:ceiling_SKATE}. The same weight of $0.5$ kg was suspended to indicate the direction of gravity. The entire mass acts to pull the gripper off from the rail. The nominal normal force per fingertip was $70$ N. Higher normal force was achievable by increasing gripper linear actuator current limits at the cost of motor duty ratio. It would be a viable option to increase grasping force for a short duration to prevent slips. In all of the climbing experiments, we only used nominal grasping force. Due to finger spine effects, the GOAT gripper stably supported SCALER on the ceiling with two grippers grasping.

\subsection{Trotting Velocity on the Ground}
SCALER 3-DoF Walking configuration was used to demonstrate SCALER is capable of trotting at a decent velocity compared to recent ground quadruped robots. SCALER trotted at normalized speed of $1.87$ /s, or $0.56$ m/s. Trot velocity was recorded as the time to pass the start and the goal line based on the video frame timestamp.

\subsection{Trotting with Payload on the Ground}
We installed a payload double SCALER's weight to compare SCALER load capacity to other quadruped robots. A weight of $3.4$ kg was attached to the back of the body, and $11.3$ kg to the top, which resulted in the normalized payload of $2.33$. SCALER trotting velocity with $14.7$ kg was at normalized speed of $0.33$ /s, or $0.13$ m/s. This payload was the absolute maximum SCALER can do trotting. The payload of $10.2$ kg was more practical and stable at $0.83$ /s or $0.25$ m/s. A mechanical efficiency metric, normalized workload capacity, reached $3.88$.

\section{CONCLUSION and FUTURE WORK}
SCALER is a true versatile quadruped robot that is capable of climbing vertical and inverted walls under Earth's gravity and running on the ground at comparable normalized speed to existing state-of-the-art robots. SCALER has shown one of the best mechanical efficiency as a limbed robot.

Although the current controllers combined with the state estimation are sufficient to conduct a performance evaluation, a more sophisticated system is necessary for fully autonomous operations. Whole-body control with gravity compensation and end-effector position control in operational space are required for a more general solution as the current feedforward controller must be tuned based on the incline.
The SCALER planner has to consider highly nonlinear kinematics, dynamics, and contact forces simultaneously to successfully perform locomotion, manipulation, and grasping in climbing. Free-climbing involves discrete decision-making since the robot must choose which holds to grasp with which leg. Solving mixed-integer nonlinear problems is challenging as the complexity exponentially increases with the number of bouldering holds in the environment. The SCALER body-translation mechanism adds another decision variable since the robot can shift workspace and move end effector positions to reach holds.


\section{Acknowledgements}
We would like to thank Varit Vichathorn, Jane Liu, Kyle Gillespie, Kate Hsieh, Julia Yuan, Jackson Kwang Hui ng, Feng Xu, for the support on SCALER  development.


\begin{thebibliography}{99}

\bibitem{free_climb} T. Bretl, "Motion Planning of Multi-Limbed Robots Subject to Equilibrium Constraints: The Free-Climbing Robot Problem," \emph{Int. J. Rob. Res.}, vol. 25, no. 4, pp. 317-342, 2006.

\bibitem{lemur3} A. Parness et al., "LEMUR 3: A limbed climbing robot for extreme terrain mobility in space," in \emph{Proc. 2017 IEEE Int. Conf. Robot. Automat.}, pp. 5467-5473, 2017.

\bibitem{anymal} M. Hutter et al., "ANYmal - a highly mobile and dynamic quadrupedal robot," in \emph{2016 IEEE/RSJ Int. Conf. Intell. Rob. Syst.}, pp. 38-44, 2016.

\bibitem{ladder_climb} M. Kanazawa et al., "Robust vertical ladder climbing and transitioning between ladder and catwalk for humanoid robots," in \emph{Proc. 2015 IEEE/RSJ Int. Conf. on Intell. Rob. Syst.}, pp. 2202-2209, 2015.

\bibitem{wheel_climbing} Y. Liu, S. Sun, X. Wu, and T. Mei, "A Wheeled Wall-Climbing Robot with Bio-Inspired Spine Mechanisms,"  \emph{Journal of Bionic Engineering}, vol. 12, no. 1, pp. 17-28, 2015.

\bibitem{rise_bd} A. Saunders et al., "The RiSE climbing robot: body and leg design",   \emph{Unmanned Systems Technology VIII} Proc. vol. 6230, 623017, 2006.


\bibitem{gekko_robot} S. Kim, M. Spenko, S. Trujillo, B. Heyneman, D. Santos, and M. Cutkosky, “Smooth vertical surface climbing with directional adhesion,” \emph{IEEE Trans. Robot.}, vol. 24, no. 1, pp. 65–74, 2008.

\bibitem{inchworm} M. Su et al., "Climbot-$\Omega$: A Soft Robot with Novel Grippers and Rigid-compliantly Constrained Body for Climbing on Various Poles," in \emph{2021 IEEE/RSJ Int. Conf. Intell. Rob. Syst.}, pp. 4975-4981, 2021.

\bibitem{octopus} Y. Sakuhara, H. Shimizu and K. Ito, "Climbing Soft Robot Inspired by Octopus," in \emph{2020 IEEE  Int. Conf. Intel. Syst.}, pp. 463-468, 2020.

\bibitem{suspension_bridge} K. H. Cho et al., "Inspection Robot for Hanger Cable of Suspension Bridge: Mechanism Design and Analysis," \emph{IEEE/ASME Trans. Mechatronics}, vol. 18, no. 6, pp. 1665-1674, 2013.

\bibitem{hubrobo} K. Nagaoka et al., "Passive Spine Gripper for Free-Climbing Robot in Extreme Terrain," \emph{IEEE Robot. and Automat. Lett.}, vol. 3, pp. 1765-1770, 2018.

\bibitem{cheetah} S. Seok  et al., "Design principles for highly efficient quadrupeds and implementation on the MIT Cheetah robot," in \emph{Proc. 2013 IEEE Int. Conf. Robot. Automat.}, pp. 3307-3312, 2013.

\bibitem{spine_horizontal}K. Weinmeister, P. Eckert, H. Witte and A. -. Ijspeert, "Cheetah-cub-S: Steering of a quadruped robot using trunk motion," in \emph{Proc. 2015 IEEE Int. Symp. Safe., Secu. Resc. Robot.},  pp. 1-6, 2015.


\bibitem{slalom} W. Haomachai et al., "Lateral Undulation of the Bendable Body of a Gecko-Inspired Robot for Energy-Efficient Inclined Surface Climbing," \emph{IEEE Robot. Automat. Lett.}, vol. 6, no. 4, pp. 7917-7924, 2021.

\bibitem{five_bar_leg} G. Kenneally, A. De and D. E. Koditschek, "Design Principles for a Family of Direct-Drive Legged Robots," \emph{IEEE Robot. and Automat. Lett.}, vol. 1, no. 2, pp. 900-907 2016.

\bibitem{stanford_doggo} N. Kau, A. Schultz, N. Ferrante and P. Slade, "Stanford Doggo: An Open-Source, Quasi-Direct-Drive Quadruped," in \emph{Proc. 2019 IEEE Int. Conf. Robot. Automat.}, pp. 6309-6315, 2019.

\bibitem{bobcat} M. P. Austin et al., "Leg Design to Enable Dynamic Running and Climbing on BOBCAT," in \emph{Proc. 2018 IEEE Int. Conf. on Intell. Rob. Syst.}, pp. 3799-3806, 2018.

\bibitem{magnetic_gripper} M. Adinehvand et al., "BogieBot: A Climbing Robot in Cluttered Confined Space of Bogies with Ferrous Metal Surfaces," in \emph{Proc. 2021 IEEE/RSJ Int. Conf. on Intell. Rob. Syst.}, pp. 2459-2466, 2021.

\bibitem{suction_gripper} Y. Yoshida and S. Ma, "Design of a wall-climbing robot with passive suction cups," in \emph{Proc. 2010 IEEE Int. Conf. Robot. Biomimetics}, pp. 1513-1518, 2010.

\bibitem{multi-surface} X. Lin, H. Krishnan, Y. Su and D. W. Hong, "Multi-Limbed Robot Vertical Two Wall Climbing Based on Static Indeterminacy Modeling and Feasibility Region Analysis," in \emph{IEEE/RSJ Int. Conf. on Intell. Rob. Syst.}, pp. 4355-4362, 2018.

\bibitem{spinybot} Sangbae Kim, A. T. Asbeck, M. R. Cutkosky and W. R. Provancher, "SpinybotII: climbing hard walls with compliant microspines," in \emph{Proc. 2005 IEEE Int. Conf. Robot. Automat.}, pp. 601-606, 2005.

\bibitem{clash} P. Birkmeyer, A. G. Gillies and R. S. Fearing, "CLASH: Climbing vertical loose cloth," \emph{Proc. 2011 IEEE Int. Conf. on Intell. Rob. Syst.} pp. 5087-5093, 2011.

\bibitem{risk_aware} Y. Shirai, X. Lin, Y. Tanaka, A. Mehta, and D. Hong, "Risk-aware motion planning for a limbed robot with stochastic gripping forces using nonlinear programming," \emph{IEEE Robot. and Automat. Lett.}, vol. 5, no. 4, pp. 4994-5001, 2020. 

\bibitem{spiny_hand} S. Wang et al., "SpinyHand: Contact Load Sharing for a Human-Scale Climbing Robot.," \emph{ASME. J. Mechanisms Robotics} Vol. 11, 2019.


\bibitem{GOAT} Y. Tanaka et al., "An Under-Actuated Whippletree Mechanism Gripper based on Multi-Objective Design Optimization with Auto-Tuned Weights," in \emph{Proc. IEEE/RSJ Int. Conf. on Intell. Rob. Syst.}, pp. 6139-6146, 2021.

\bibitem{spine_cell} W. Shiquan, H. Jiang, and M. R, Cutkosky, “Design and Modeling of Linearly-Constrained Compliant Spines for Human-Scale Locomotion on Rocky Surfaces.” \emph{Int. J. Rob. Res.}, vol. 36, no. 9, 2017.


\bibitem{alex_admittance} A. Schperberg et al., "Auto-Calibrating Admittance Controller for Robust Motion of Robotic Systems," preprint arXiv:2207.01033, 2022.

\bibitem{admittance} C. Ott, R. Mukherjee and Y. Nakamura, "Unified Impedance and Admittance Control," in \emph{Proc. 2010 IEEE Int. Conf. Robot. Automat.}, pp. 554-561, 2010.

\bibitem{estimation} Bloesch, Michael, et al. "State estimation for legged robots-consistent fusion of leg kinematics and IMU.", in \emph{Proc. 2013 R:SS}, Robotics 17 (2013): 17-24, 2013.

\bibitem{MPC} J. Di Carlo, P. M. Wensing, B. Katz, G. Bledt and S. Kim, "Dynamic Locomotion in the MIT Cheetah 3 Through Convex Model-Predictive Control," in \emph{IEEE/RSJ Int. Conf. on Intell. Rob. Syst.}, pp. 1-9, 2018.

\bibitem{contact_rich} Y. Shirai et al., "Simultaneous Contact-Rich Grasping and Locomotion via Distributed Optimization Enabling Free-Climbing for Multi-Limbed Robots," \emph{2022 IEEE/RSJ Int. Conf. on Intell. Rob. Syst.}, 2022. 

\bibitem{quad_comparison} P. Biswal, P. K. Mohanty, "Development of quadruped walking robots: A review,"  \emph{Ain Shams Engineering Journal}, vol. 12, no. 2, pp. 2017-2031, 2021.


\end{thebibliography}
\end{document}